\documentclass[10pt, a4paper]{article}

\usepackage{lrec-coling2024} 
\usepackage{booktabs,multirow,xcolor,amsmath,amssymb,subcaption}

\title{MSNER: A Multilingual Speech Dataset for Named Entity Recognition}

\name{Quentin Meeus$^{1,2}$, Marie-Francine Moens$^{1}$, Hugo Van hamme$^{2}$} 

\address{
20th Joint ACL-ISO Workshop on Interoperable Semantic Annotation \\
$^{1}$ LIIR Lab, Computer Science Dpt., KU Leuven \quad 
$^{2}$ PSI, Electrical Engineering Dpt., KU Leuven\\
Quentin.Meeus@kuleuven.be\\
}

\abstract{
While extensively explored in text-based tasks, Named Entity Recognition (NER) remains largely neglected in spoken language understanding. Existing resources are limited to a single, English-only dataset. This paper addresses this gap by introducing MSNER, a freely available, multilingual speech corpus annotated with named entities. It provides annotations to the VoxPopuli dataset in four languages (Dutch, French, German, and Spanish). We have also releasing an efficient annotation tool that leverages automatic pre-annotations for faster manual refinement. This results in 590 and 15 hours of silver-annotated speech for training and validation, alongside a 17-hour, manually-annotated evaluation set. We further provide an analysis comparing silver and gold annotations. Finally, we present baseline NER models to stimulate further research on this newly available dataset.\\ \newline 
}

\begin{document}

\maketitleabstract

\section{Introduction}
\label{sec:intro}
In an increasingly interconnected world where language knows no boundaries, the field of Speech Processing is undergoing a transformative shift towards multilingual applications. One such pivotal area is Spoken Named Entity Recognition (Spoken NER). Named Entity Recognition (NER) is a natural language processing (NLP) task that involves the identification and categorization of named entities within a text, typically into predefined categories such as names of persons, organizations, locations, dates, numerical values, and more. The primary objective of NER is to automatically recognize and extract specific pieces of information from unstructured text, making it easier to analyze and understand the content. NER plays a crucial role in various NLP applications, including information retrieval, question answering, sentiment analysis, and language understanding. In contrast, \textit{Spoken NER} extracts named entities from audio documents, a task that is considerably more challenging. Indeed, aside from the inherent difficulties associated with speech processing, Spoken NER requires not only to identify and classify the entities, but also to transcribe them correctly. Variability in pronunciation, accents, and dialects can make the detection and especially the spelling of named entities very challenging. On the other hand, prosody, intonation and emphasis are cues that may be crucial for NER but are not readily available in written text. Recognizing the pressing need to facilitate cross-lingual research and to provide comprehensive evaluation resources for Spoken NER models, we have undertaken the task of manually annotating the popular speech dataset VoxPopuli's test sets in four languages: Dutch, French, German, and Spanish. Additionally, we also provide machine-made annotations on the training and validation sets.

In the following sections, we provide a detailed overview of our efforts in the domain of Spoken NER. First, we give an overview of related works and datasets. Then, we introduce the newly annotated dataset and provide information about its size, multilingual coverage, and its potential significance in advancing Spoken NER technology. Additionally, we describe the methodology employed in the dataset's creation, breaking down the annotation process and data preparation. We also introduce the user-friendly annotation interface we've developed for this purpose. Furthermore, we present the results of various experiments and benchmarks conducted using this dataset. These experiments demonstrate its utility in evaluating Spoken NER models across the chosen languages, highlighting its role in advancing research and development in this field.

In summary, this article describes our contributions to the field of multilingual Spoken NER, including the dataset's creation, annotation methodology, and its role in advancing research in this domain.

\section{Literature Review}
In the field of NLP, there is not one unified label set. Both generic and specialized datasets exist with their own label sets defined. Specialized datasets might cover large amounts of topics with specific vocabulary and entities. For example, a NER system for doctors would include medications, dosages, medical reasons, etc. \cite{i2b2}, and biomedical entities include names of proteins, chemical, disease, or species \cite{bio-ner}. Other datasets provide more generic entities that cover broader landscapes. One of the most widely used is CoNLL-2003 \cite{conll}, although it comes with only four entity types (LOC, ORG, PER and MISC). OntoNotes v5 enriches this set with 14 more classes (Table \ref{tab:ner-labelset}), to include things such as numbers, dates, and laws. Its high quality makes it one of the most widely used NER datasets, although it only covers three languages: English, Arabic and Chinese. Another notable mention is \citet{tedeschi2021}, which adds a few more generic classes to OntoNotes definitions to cover things such as animal names, diseases, food, and plants, and released a dataset derived from Wikipedia where named entities were annotated automatically with an annotation pipeline that effectively combined pretrained language models and knowledge-based approaches. A follow-up dataset was published covering more languages \cite{multinerd}.

Currently, we know of only one Spoken NER dataset that is openly distributed as SLUE \cite{SLUE}. This is an annotated subset of the larger VoxPopuli dataset \cite{voxpopuli}, which comprises audio recordings and corresponding transcripts of sessions held in the European Parliament. The annotated portion of the dataset include approximately 25 hours of speech, divided into three subsets: 3/5 for training, 1/5 for validation, and 1/5 for testing purposes. While this initiative is a significant step forward, SLUE exclusively covers the English language. They used the same entities as OntoNotes \citeplanguageresource{ontonotes} although in practice, they combine some types and remove rare ones to produce a new label set (Table \ref{tab:ner-labelset}, Column 2). 

Another task in spoken language understanding is similar to Spoken NER: slot filling. This is the identification of information relevant to specific applications, such as flight booking \cite{atis}. Although they share many grounds, there is a major difference: slot filling relates to a specific application, and in this regard, covers a much narrower domain than NER, often consisting of short commands for a computer interface \citep{fluent,snips,slurp,timers-and-such,grabo} or a booking system \citep{atis}.

Since the vast majority of entity recognition datasets are text-based, the same goes for the applications. Consequently, NER is often framed as a token classification task, where each word or word piece must be assigned an entity type. Since an entity can cover many tokens, the entity classes are redefined in the BIO format, a widely used tagging scheme in NER tasks \cite{ramshaw-marcus-1995}. This format provides a structured way to label and distinguish the boundaries of named entities within the text. Each word or token is tagged with one of three labels: ``B'' marks the beginning, or first word of an entity, ``I'' indicates the continuation of the named entity and always follows the ``B'' tag, and ``O'' is used for words that are not part of an entity. This marker, together with the entity type, makes the target for the classification task. Other annotation schemes are extensions of this (e.g. IO, IOBES, IOE, etc.). The major drawback of the BIO format is its inability to represent nested entities.

The modern approach to NER is to add linear layers to a pretrained language model and fine-tune it on the chosen NER dataset. Sometimes, a conditional random field (CRF) \cite{CRF} is added to learn the transition probabilities between the label classes \cite{TNER}. In Spoken NER, the two main approaches are pipeline and end-to-end models. As the name suggests, pipeline models first use automatic speech recognition to transcribe an audio recording, then use NER to predict the entities. In contrast, end-to-end models do not force the model to make hard decisions by choosing one token over another. Instead, it predicts entities directly from the hidden states. Finally, hybrid models or multitask models predict both the entities and the transcriptions simultaneously \cite{Meeus2023}.

\begin{table}[h]
    \footnotesize
    \centering
    \begin{tabular}{lllll}
        \toprule
        subset & language & duration & size & entities \\
        \midrule
        \multirow{4}{*}{train} & DE &  224.5 h &  86,410 & 97,492 \\
            & ES &  141.5 h &  47,611 & 66,482 \\
            & FR &  186h &  65,952 & 80,255 \\
            & NL &  38.5 h &  16,533 & 19,566 \\
        \hline
        \multirow{4}{*}{dev} & DE & 4h &  1,610 & 1,880 \\
            & ES & 4h48 &  1,529 & 2,094\\
            & FR & 4h22 &  1,527 & 1,884 \\
            & NL & 2h16 &  963 & 1,074 \\
        \hline
        \multirow{4}{*}{test} & DE &  5h &  1,966 & 2,061 \\
            & ES &  5h &  1,512 & 2,198 \\
            & FR &  4h30 &  1,656 & 2,004 \\
            & NL &  2h30 &  1,120 & 1,272 \\
        \bottomrule
    \end{tabular}
    \caption{MSNER Dataset statistics}
    \label{tab:stats}
\end{table}

\begin{table*}[ht]
\footnotesize
\centering
\begin{tabular}{lllrrrrl}
\toprule
OntoNotes5 & SLUE & DE & ES & FR & NL & Examples \\
\midrule
date & \multirow{2}{*}{WHEN} & 307 & 276 & 243 & 113 & 125 years ago, 15 maart, 1815—1830, 1997 \\
time &  & 12 & 21 & 10 & 8 & 24 hours, acht uur, de hele dag, mañana \\
\hline
cardinal number & \multirow{5}{*}{QUANT} & 136 & 167 & 123 & 91 & 1, 10, 10 miljoen, 11, 11 billion \\
ordinal number &  & 82 & 100 & 79 & 45 & First, Ten derde, dritten \\
quantity &  & 6 & 2 & 5 & 1 & one and a half meter, two inches \\
money &  & 26 & 16 & 18 & 8 & 200 million EUR, Dertig miljoen euro \\
percent &  & 21 & 28 & 13 & 22 & 1 procent, 100\%, 15 Prozent \\
\hline
geopolitical area & \multirow{2}{*}{PLACE} & 259 & 285 & 283 & 176 & Amsterdam, Australië, Barcelona, Belgium \\
location &  & 128 & 139 & 214 & 110 & Afrika, Balkanlanden, Europe \\
\hline
group & NORP & 229 & 244 & 285 & 213 & African, American, Christian \\
organization & ORG & 621 & 638 & 527 & 362 & Amnesty International, Charlie Hebdo \\
law & LAW & 64 & 108 & 33 & 22 & Paris Accords, US Constitution \\
person & PERSON & 123 & 131 & 100 & 67 & Angela Merkel, Barroso, Beyoncé \\
facility & - & 6 & 2 & 8 & 12 & Guantánamo, White House \\
event & - & 23 & 25 & 21 & 8 & Europees Semester, Rio conferentie \\
work of art & - & 6 & 3 & 4 & 4 & Green Book, Koran \\
product & - & 4 & 1 & 2 & 8 & 2G, 4G, 5G, iPhone \\
language & - & 3 & 12 & 6 & 2 & Latin, Nederlands, Español \\
\bottomrule
\end{tabular}
\caption{Number of annotated entities per entity type in the test sets. Column SLUE correspond to the `combined' entity set proposed by \citet{SLUE}.}
\label{tab:ner-labelset}
\end{table*}

\section{Dataset description}
\label{sec:dataset}
\begin{figure}[h]
    \footnotesize
    \centering
    \includegraphics[width=\linewidth]{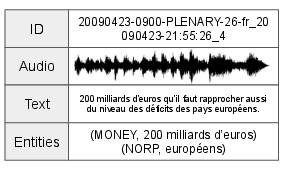}
    \caption{Annotated example}
    \label{fig:annotation}
\end{figure}

The MSNER dataset is an annotated version of the VoxPopuli dataset \cite{voxpopuli} in four languages -- Dutch, French, German, and Spanish. VoxPopuli is a collection of recorded sessions from the European Parliament, segmented to contain one or more sentence by one speaker. For each language in scope, we provide three annotated subsets (Table \ref{tab:stats}): a training and development set with machine-generated ``silver'' annotations, and a test set with manual ``gold'' annotations. The subsets of the four languages in scope were annotated according to OntoNotes' 18 classes. The test sets were manually annotated by the authors following the methodology outlined in Section \ref{sec:methodo}. Each example in the annotated dataset contains the VoxPopuli ID to identify the relevant audio recording in the original dataset, the transcribed sentence and the annotated named entities, that is, the list of entities, each composed of a text and a label component (Figure \ref{fig:annotation}). For the silver label datasets, we also provide a probability score of each predicted entity. We discuss in Section \ref{sec:experiments} how this number is related to the uncertainty of the model.

We use the 18-classes OntoNotes label set \citeplanguageresource{ontonotes}. However, following the example from \citet{SLUE}, we provide annotations by using an alternative label set that combines entity types like places or numbers and discard the rarest classes like languages, events, and work of art (Table \ref{tab:ner-labelset} Column 2). 

\section{Methodology}
\label{sec:methodo}
\begin{figure*}[ht]
    \centering
    \includegraphics[width=\textwidth]{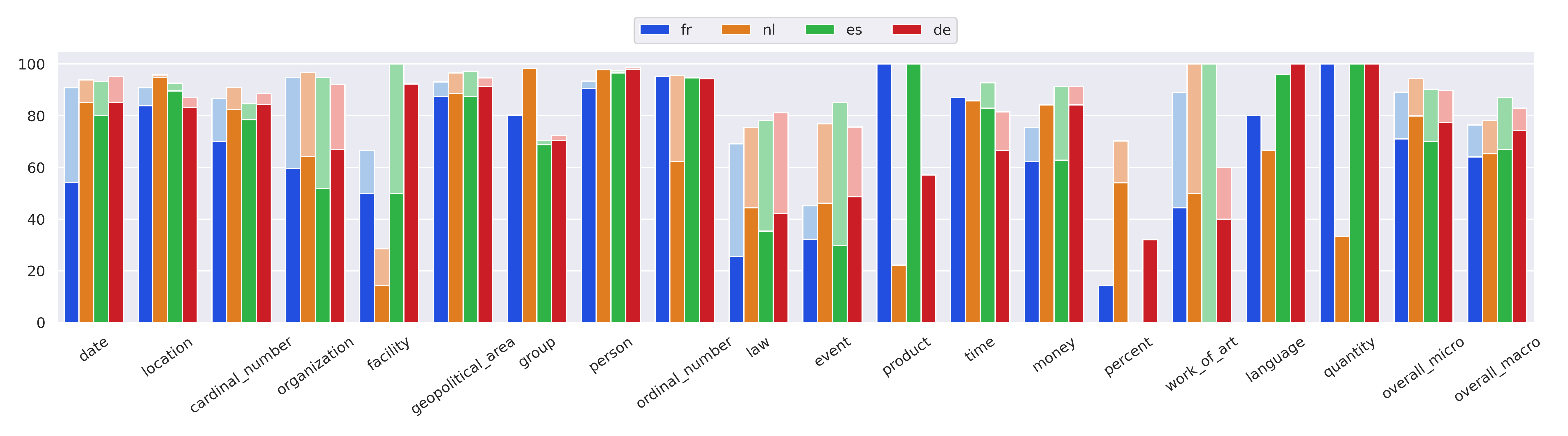}
    \vspace{-24pt}
    \caption{Evaluation of text-based pretrained NER model against our annotations. Bright colors correspond to the F1-score and faded colors correspond to the label-F1 score, a metric that ignores spelling mistakes and segmentation errors.}
    \label{fig:breakdown}
\end{figure*}
We provide two kinds of label quality: machine-generated ``silver'' labels and human-annotated ``gold'' labels. For obvious reasons, the silver labels are much cheaper and easier to produce. Therefore, we only provide human-made annotations for the test sets, and the training and validation sets annotations are entirely machine-generated. The methodology follows these four broad steps: (1) filtering out recordings without or with misaligned transcripts, (2) generate silver labels for all subsets, (3) manually annotate the test sets and (4) verify the human-made annotations to identify and rectify potential labelling errors. We detail each step in the following paragraphs.

\subsection{Filtering}
The VoxPopuli dataset contains a few alignment errors between the spoken content and its corresponding transcript. To address this issue, we employed an automatic speech recognition (ASR) system, initially transcribing the spoken utterances and subsequently calculating the word error rate by comparing the ASR-generated sentence to the provided transcript. For this task, we opted for the Whisper large v2 ASR model \citep{whisper}, because it showed near state-of-the-art performance across the selected languages. Notably, this model has been meticulously trained on extensive, well-curated data to perform both audio translation and transcription tasks.\\
For the training and development sets, we filter out examples with a WER larger than 20\%, without verifying that the excluded examples were indeed problematic. This discards about 20\% of the German and Dutch utterances, 10\% of the French examples and 6\% of the Spanish utterances.\\
For the test sets, instances where the word error rate (WER) between the machine-generated transcription and the original transcript exceeded 20\%, we conducted a meticulous review process. This involved listening to the audio recording and cross-referencing it with the existing transcript. When feasible, we made necessary corrections to the transcript. However, in cases where multiple speakers were heard in the recording or no speech is present, we removed the problematic utterance from the dataset.

\subsection{Pseudo-annotations}
We employed an established text-based Named Entity Recognition (NER) model to predict entities within the gold transcript. We chose to use the XLM-RoBERTa large pretrained model \citep{xlm}, fine-tuned specifically on the OntoNotes v5 dataset \citeplanguageresource{ontonotes}. This model is readily accessible through the HuggingFace repository\footnote{\url{https://huggingface.co/asahi417/tner-xlm-roberta-base-ontonotes5}}.\\ While it's important to note that this particular model's fine-tuning was conducted solely on English data, its robustness and efficacy across multiple languages were remarkable. In our evaluation, we observed impressive performance, with most sentences annotated correctly.

\subsection{Annotation Tool}
For each of the 6,254 pre-annotated sentences in the test sets, we corrected the annotations predicted by the model. For this purpose, we have developed a command line tool to quickly add, edit, merge or remove annotations in a sentence. This utility displays the pre-annotated sentence with a summary of the annotations below. Annotations appear as colored XML tags both in the text and in the summary. An annotated English translation can be displayed. The annotator then has access to both the original sentence and the translation to make sure that the annotations are as accurate as possible. When presented with a sentence, the annotator has the choice to add a new annotation, delete an existing one, merge two annotations together or modify an annotation, either by changing the type or by adding or removing words. Once a sentence has been annotated, it is saved to a file in JSON format. Following this methodology and with the help of this tool, we were able to save a lot of time and effort without sacrificing accuracy. For this reason, we make the tool available online so that others will have the opportunity to contribute to this field of research by easily annotating more data in many more languages.

As mentioned in Section \ref{sec:dataset}, we not only provide annotations according to OntoNotes 18 classes, but also the 7-classes combined set proposed in \citet{SLUE}. However, we chose to completely reannotate the examples where entities are removed, instead of simply removing all the annotations of the same type from the dataset. To illustrate this, consider the following example:
\begin{quote}
\verb|<event>| 15th conference on speech of Toronto \verb|</event>|
\end{quote}
According to the combined set conversion rules (Table \ref{tab:ner-labelset}), all the entities of type \verb|<event>| are to be discarded. Doing that would lead to two unannotated entities, `15th' as a number and `Toronto' as a place. Instead, we re-annotate the examples containing removed entities to make sure that we are not penalizing the models for correct assumptions.

\begin{figure*}[ht]
    \centering
    \includegraphics[width=\textwidth]{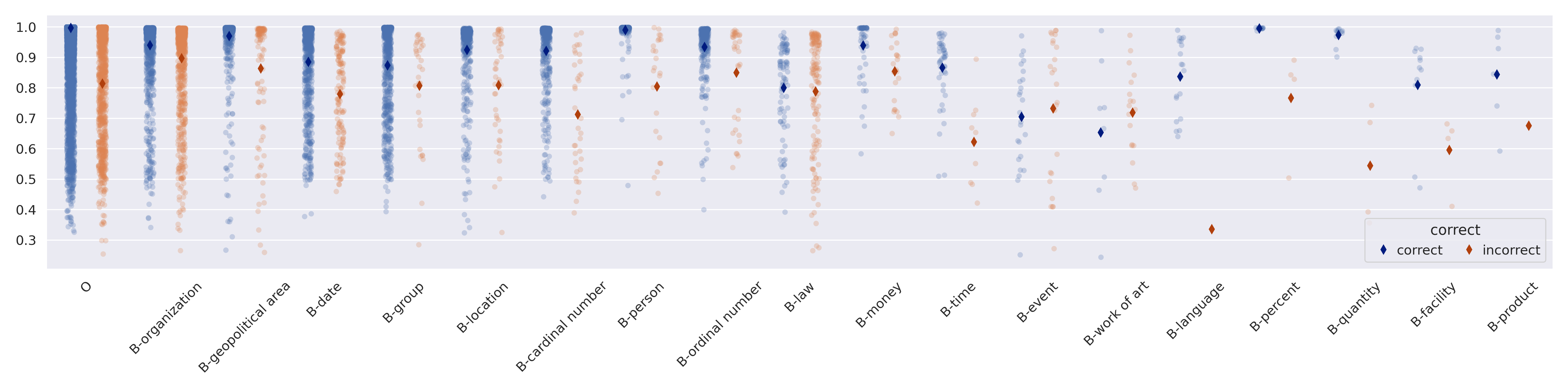}
    \vspace{-18pt}
    \caption{Distribution of predicted probability score per class given the target class for the text-based model's predictions}
    \label{fig:conditional-mean}
\end{figure*}

\begin{figure}[ht]    
    \centering
    \includegraphics[width=\linewidth]{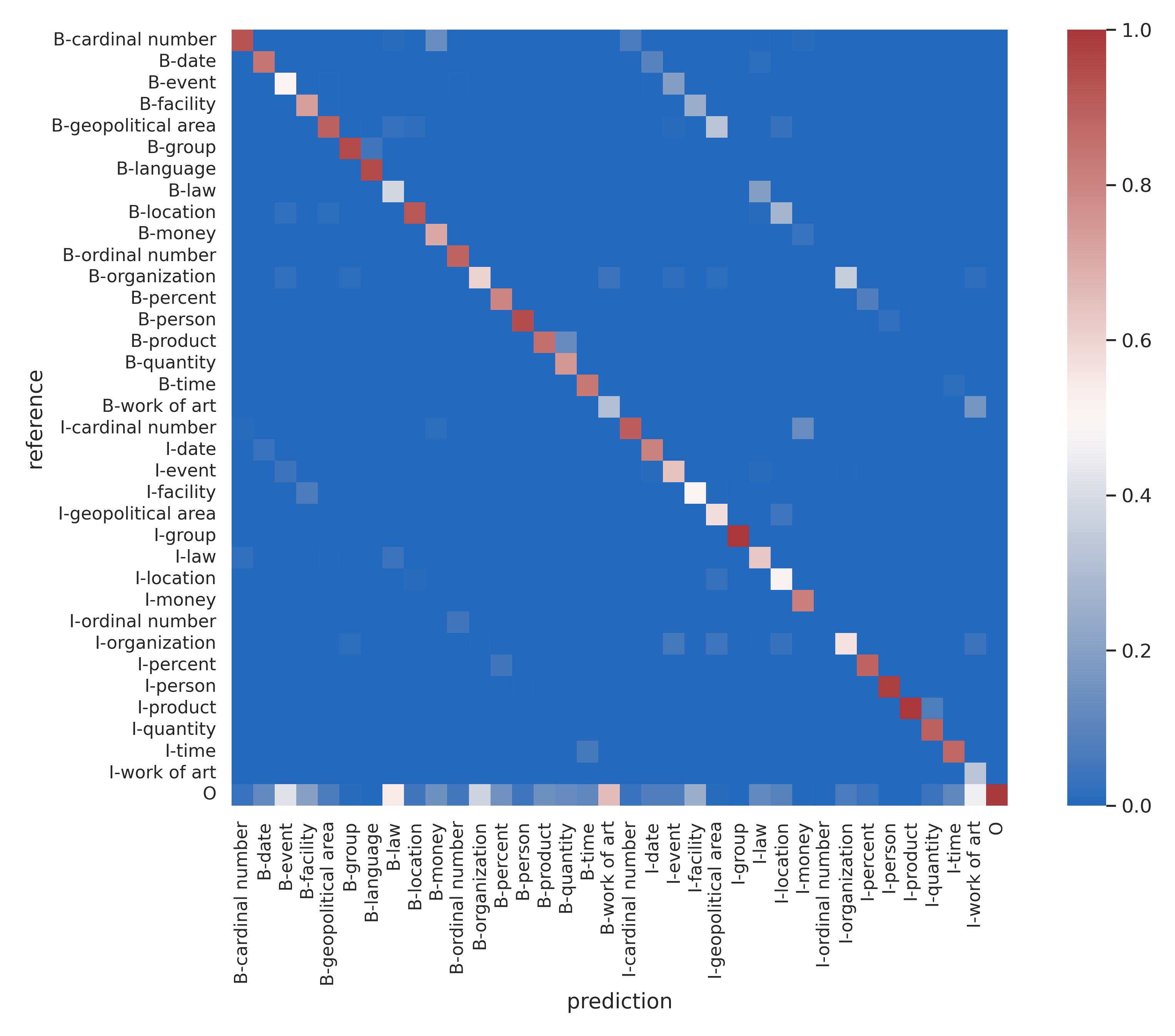}
    \caption{Confusion matrix, normalized to show the probability distribution of the tags predicted with the text-based model.}
    \label{fig:confusion-matrix}
\end{figure}

\subsection{Verification}
Finally, we verify the integrity of the test annotations by deriving a number of heuristics and rules that the annotations must abide. This involved grouping the annotations by category and verify each list one by one, comparing them to one another, searching in the text for frequent annotated terms to identify missing annotations, etc. In this last step, we also fix some remaining transcription issues. For example, we realized that VoxPopuli transcripts omitted the symbol ``\%'', and sometimes the word ``thousands'' (in all languages). Consequently, for all entities marked as cardinal number, we added the missing tokens when necessary, following the rules specific to the language\footnote{In French and in Spanish, the symbol ``\%'' is generally used, but in German and in Dutch, the word is more commonly spelled as Prozent or procent, respectively.}. Another error often made by the text-based NER model is to predict the article as being part of the entity. As multiple sources advocate against doing so, we abided by the main guidelines \cite{OntoNotes-Guidelines,benikova-etal-2014-nosta}.

\subsection{Distribution}
\label{sec:data:dist}
The annotated datasets are distributed in two formats: As JSON Lines files available on GitHub\footnote{\url{https://github.com/qmeeus/MSNER}}, and on the HuggingFace repository \cite{huggingface}. There is one file per subset and per language, where each line is an annotated example. The audio files can be obtained by downloading VoxPopuli and matching the audio ID. The dataset version hosted on HuggingFace contains the audio recordings and the preprocessed annotations in BIO format, so that a researcher can already use the dataset after only two lines of code.

\section{Evaluation Metrics}
Following \citet{SLUE}, we recommend evaluating model predictions with the micro-averaged F1-score. The F1-score is the harmonic mean of precision and recall, calculated from an unordered list of named entities predicted for each utterance. Precision is the proportion of correctly predicted entities among all predicted entities, and recall is the proportion of ground truth entities that were correctly detected. An entity is considered to be predicted correctly if both the type and spelling are identical to the ground truth. To allow multiple entities with the same spelling and type in a sentence, we add a unique identifier to each entity/type pair. We recommend using the micro-averaged F1-score because the dataset is unbalanced. The label F1-score only considers the predicted type of the entity for correctness, leaving the transcribed entity out of the computations. This metric ignores spelling mistakes and segmentation errors.   
We provide an evaluation script\footnote{\url{https://raw.githubusercontent.com/qmeeus/MSNER/main/src/evaluate.py}} to compute these metrics and generate a breakdown of the prediction results per entity type.

\section{Experiments}
\label{sec:experiments}
\subsection{Setup}
The first analysis compares the annotated test sets to the pseudo-annotations generated by the text-based NER model. Since the silver-label training and validation sets were generated with this model, this analysis is valuable for anyone intending to use these datasets for training. Indeed, it gives insights into the entities that are often confused with one another or remain undetected. It also gives some insights on the reliability of the model's confidence score in assessing whether a prediction is correct.

We also consider two methods to predict named entities from speech, with a pipeline and an end-to-end model. The end-to-end model is a transformer encoder-decoder trained to perform both ASR and NER with a multitask objective \cite{Meeus2023}. This model is initialized from Whisper Large V2 \cite{whisper}, with an additional SLU module connected to the layers of the decoder with an adaptor. The end-to-end model was fine-tuned on English SLUE-VoxPopuli \cite{SLUE}. The pipeline model transcribes the audio files and subsequently annotates the transcriptions. For the ASR model, we use Whisper Large V2 \cite{whisper}. For the pipeline model, we provide two options to allow for a better comparison. In Table \ref{tab:text-vs-pipe}, we use XML-RoBERTa fine-tuned on OntoNotes v5 \citeplanguageresource{ontonotes} and compare it to the predictions generated by the text-based NER model from the gold transcripts. In Table \ref{tab:pipe-vs-e2e}, we fine-tuned the same XML-RoBERTa on SLUE-VoxPopuli \cite{SLUE}, which provides a fair comparison to the end-to-end model.

Although both models rely on multilingual pretrained models, the fine-tuning dataset is entirely in English. Therefore, we evaluate the ability of these models to generalize from one language (English) to other languages (Dutch, French, German, and Spanish). Before computing the F1-scores, we normalize the text by putting it in lower case and removing symbols. It should be noted that the evaluation script does normalize the text further, which could have its importance depending on the model to be evaluated.

All results are presented on the human-annotated test sets proposed in this article.

\subsection{Results}
Figure \ref{fig:conditional-mean} shows the distribution of calculated probabilities for predicted `B' and `O' tags conditional to whether they were predicted correctly or not. For each token position $k$, the probability of the most likely tag $i^*$ is computed as follows: 
$$P(y^k = i^*) = \max_{i}{\frac{e^{z^k_i}}{\sum_{j} e^{z^k_{j}}}}$$ 
where $z^k_{1..N}$ are the logits predicted by the model for the token at position $k$. We observe that, on average, annotations for which there was no agreement between the annotator and the NER model were predicted with a lower probability than annotations that were correctly annotated from the start. However, we observe major differences between the class distributions. For the most frequent classes, like `O', `organization' or `date', the probability distributions overlap considerably, and 
one should be careful if using this score as a proxy for the model's uncertainty. This is not surprising, as transformers are known to be overconfident \cite{pmlr-v216-ye23a}. For rare quantitative classes like `percent' and `quantity', the model shows confidence when predictions are correct, and uncertain otherwise. This indicates that for those particular classes, the given probability could be relied upon when estimating the model's uncertainty. The score breakdown by entity and language (Figure \ref{fig:breakdown}) indicates that in general, there are no major differences across languages, except for rare classes, where the variability increases significantly.

Figure \ref{fig:confusion-matrix} shows the confusion matrix of the NER model predictions against the manual annotations. Most errors are undetected entities (bottom row in Figure \ref{fig:confusion-matrix}) and segmentation errors (I-tags predicted instead of B-tags and inversely, are visible on the lighter diagonals above and below the main diagonal). Some entities remain undetected more often than not, e.g. ``work of art'' and ``event'', which is a sign that predictions are less reliable for these rare classes. Some other types are often confused with one another, like ``money'' and ``cardinal number''. However, all types seem to have at most two confused types. We notice that ``geopolitical area'' is most often confused with ``location'' and ``law''. In the latter case, this is because many laws are named after cities (e.g. the Paris Agreement, the Warsaw Treaty). 

Table \ref{tab:text-vs-pipe} compares the text-based NER predictions with the NER predictions obtained from the ASR transcript and generated by the same text-based NER model. The OntoNotes dataset, although in English, provides many well-curated annotations and the NER model trained on this dataset seem to generalize well to the other languages. However, this model was not trained to handle automatic transcripts and we observe a considerable drop in performance when it is asked to process ASR outputs. To make a fair comparison with the end-to-end model, we fine-tune XML-RoBERTa on SLUE-VoxPopuli and report the results in Table \ref{tab:pipe-vs-e2e}. The fine-tuning dataset being of much modest size (14.5 hours of training data), the models do not have many examples to learn from. The end-to-end model has a slight advantage because it learns simultaneously the ASR and NER tasks, and it is able to share part of its architecture between both tasks. For example, it seems well able to identify the presence of entities despite a lot of transcription and segmentation errors, as evidenced by the large label F1-score. In contrast, the pipeline suffers much more from the transcription errors because it was pretrained on curated texts and is not expecting noisy ASR transcriptions.

The text-based NER model performs best for Dutch, then German, French and finally Spanish. As the model was trained on English annotations, this ranking is not a surprise, although the ability of the model to transfer to other languages is impressive. However, for the speech processing models, the same conclusion cannot be drawn.
The entity F1-score seem to be correlated with the word error rate, which is influenced by the availability of the different languages in the pretraining set. In other words, for speech models, this is the model's ability to transcribe foreign languages that will drive the quality of the predictions, rather than how similar the evaluation and the pretraining language are. The label-F1 indicates how accurate a model is at detecting the presence of entity types, disregarding of its ability to transcribe it correctly. Looking at those numbers, we observe again the same behavior as with the text-based entity predictions, namely that entities are more likely to be accurately detected when the evaluation language is more similar to the finetuning language.

\begin{table}[h]
    \centering
    \footnotesize
    \begin{tabular}{cc|cccc}
        \toprule
        Model & Metric & DE & ES & FR & NL \\
        \midrule
        \multirow{2}{*}{Gold} & F1 ($\uparrow$) & 77.4 & 70.1 & 71.1 & 79.9 \\
            & Label-F1 ($\uparrow$) & 89.7 & 90.3 & 89.1 & 94.4 \\
        \hline
        \multirow{3}{*}{ASR} & F1 ($\uparrow$) & 52.4 & 50.6 & 44.7 & 52.7 \\
            & Label-F1 ($\uparrow$) & 66.2 & 63.6 & 59.4 & 66.1 \\
            & WER ($\downarrow$) & 12.0 & 8.6 & 11.1 & 13.1 \\
        \bottomrule
    \end{tabular}
    \caption{Performance of text-based NER model trained on OntoNotes. Gold corresponds to the model's predictions from the gold transcripts and ASR corresponds to the model's predictions on the ASR transcripts.}
    \label{tab:text-vs-pipe}
\end{table}

\begin{table}[h]
    \centering
    \footnotesize
    \begin{tabular}{cc|cccc}
        \toprule
        Model & Metric & DE & ES & FR & NL \\
        \midrule
        \multirow{3}{*}{Pipeline} & F1 ($\uparrow$) & 30.8 & 36.3 & 37.2 & 36.3 \\
            & Label-F1 ($\uparrow$) & 42.7 & 51.6 & 49.5 & 45.9 \\
            & WER ($\downarrow$) & 12.0 & 8.6 & 11.1 & 13.1 \\
        \hline
        \multirow{3}{*}{End2End} & F1 ($\uparrow$) & 38.3 & 41.3 & 39.6 & 31.2 \\
            & Label-F1 ($\uparrow$) & 76.8 & 77.1 & 78.3 & 78.4 \\
            & WER ($\downarrow$) & 13.3 & 10.5 & 14.5 & 18.2 \\
        \bottomrule
    \end{tabular}
    \caption{Provided baselines on the annotated test sets for a pipeline ASR/NER model and an end-to-end multitask model. Both models were fine-tuned on SLUE-VoxPopuli \cite{SLUE}}
    \label{tab:pipe-vs-e2e}
\end{table}

\section{Conclusion}
In this manuscript, we have presented MSNER, a new dataset for evaluating multilingual Spoken NER systems. Although NER is a popular topic in NLP, this task has remained mostly unexplored in speech processing and spoken language understanding. To address this issue, we have used a pretrained model to annotate the VoxPopuli training and validation subsets in Dutch, French, German, and Spanish. Additionally, to provide researcher with a gold standard dataset for evaluating their Spoken NER models, the authors have manually annotated the test sets for these subsets. By analyzing the predictions of a text-based NER model, and comparing them with our annotations, we were able to identify points of attentions for researchers who intend to train a model on silver annotations. For example, in some cases, the model confidence on the predictions can serve as a basis to estimate the correctness of the prediction, but this must be done carefully, since we have seen that transformers can be overconfident. Counter-intuitively, we have shown that most frequent classes are not always the ones where the model's uncertainty is most reliable. We also looked at the classes that were often confused with one another, which gave us some ideas about which errors might be present in the training and validation sets. 

We also provide baselines on the newly annotated evaluation subsets. We selected a pipeline and an end-to-end SLU model, both fine-tuned on English SLUE VoxPopuli \cite{SLUE}, and we evaluate them on the manually annotated test sets. We saw that in a low resource scenario, the end-to-end model seems to benefit from learning simultaneously to transcribe and to annotate, which allows a better generalization across languages than the pipeline model fine-tuned on the same dataset. Finally, we found that the performance of text-based models on unseen languages is correlated with the similarity of the evaluation language with English. However, for speech models, this is the multilingual transcription accuracy that is the main driver for NER performance. Interestingly, we have seen that the end-to-end model was able to identify the presence of entities much better than the pipeline model, despite a similar overall performance, which illustrate the advantage of sharing parameters across tasks. 

\section{Acknowledgements}
This research received funding from the Flemish Government under the ``Onderzoeksprogramma Artificiele Intelligentie (AI) Vlaanderen'' programme.

\nocite{*}
\section{Bibliographical References}\label{sec:reference}

\bibliographystyle{lrec-coling2024-natbib}
\bibliography{references}

\section{Language Resource References}
\label{lr:ref}
\bibliographystylelanguageresource{lrec-coling2024-natbib}
\bibliographylanguageresource{languageresource}

\end{document}